\documentclass{article}

\usepackage{arxiv}

\usepackage[utf8]{inputenc} 
\usepackage[T1]{fontenc}    
\usepackage{hyperref}       
\usepackage{url}            
\usepackage{booktabs}       
\usepackage{amsfonts}       
\usepackage{nicefrac}       
\usepackage{microtype}      
\usepackage{lipsum}
\usepackage[misc]{ifsym}

\usepackage{amsmath}
\usepackage{graphicx}
\usepackage{amsmath}
\usepackage{amssymb}
\usepackage{amsthm}
\usepackage{booktabs}
\usepackage{algorithm}
\usepackage{algorithmic}
\usepackage{color}
\usepackage{multirow}
\usepackage{times}
\usepackage{soul}
\usepackage{url}

\title{Hateful Memes Detection via Complementary Visual and Linguistic Networks}

\author{Weibo Zhang\textsuperscript{1,2,*}, Guihua Liu\textsuperscript{1,2,*}, Zhuohua Li\textsuperscript{1,2}, Fuqing Zhu\textsuperscript{1,\Letter}
\\
\textsuperscript{\rm 1}Institute of Information Engineering, Chinese Academy of Sciences, Beijing, China\\
\textsuperscript{\rm 2}School of Cyber Security, University of Chinese Academy of Sciences, Beijing, China\\
\texttt{\{zhangweibo,liuguihua,lizhuohua,zhufuqing\}@iie.ac.cn}
\\
\textsuperscript{\rm * Two authors contribute equally to this work. \Letter  Corresponding author.}
}

\begin{document}
\maketitle

\begin{abstract}
Hateful memes are widespread in social media and convey negative information. The main challenge of hateful memes detection is that the expressive meaning can not be well recognized by a single modality. In order to further integrate modal information, we investigate a candidate solution based on complementary visual and linguistic network in Hateful Memes Challenge 2020\footnote{https://www.drivendata.org/competitions/64/hateful-memes/}. In this way, more comprehensive information of the multi-modality could be explored in detail. Both \emph{contextual-level and sensitive object-level information} are considered in visual and linguistic embedding to formulate the complex multi-modal scenarios. Specifically, a pre-trained classifier and object detector are utilized to obtain the contextual features and region-of-interests (\emph{RoIs}) from the input, followed by the position representation fusion for visual embedding. While linguistic embedding is composed of three components, \emph{i.e.}, the sentence words embedding, position embedding and the corresponding Spacy embedding (\emph{Sembedding}), which is a symbol represented by vocabulary extracted by Spacy. Both visual and linguistic embedding are fed into the designed Complementary Visual and Linguistic (CVL) networks to produce the prediction for hateful memes. Experimental results on Hateful Memes Challenge Dataset demonstrate that CVL provides a decent performance, and produces $78.48\%$ and $72.95\%$ on the criteria of AUROC and Accuracy. Code is available at \url{https://github.com/webYFDT/hateful}.
\end{abstract}


\section{Introduction}
Hateful memes detection is a task that gives a pair of <\emph{image and text}> to discriminate whether there are hateful memes. The widespread spread of social media makes hateful memes detection task more urgent, while manual or uni-modal can not deal with the increasing the volume and complex samples processing. Automatic multi-modal hateful memes detection is inevitable trend, which focuses on how to enrich the multi-modal information between various modals as one of the critical entry point. However, the input features are not fully exploited.

Recently, many pretrain-then-transfer Visual and Linguistic (V\&L) framework \cite{lu2019vilBERT,li2019visualBERT,sun2019videoBERT,su2019vl,tan2019lxmert} has been designed to integrate various visual-language tasks, such as \emph{Visual Question Answering} (VQA), \emph{Visual Commonsense Reasoning} (VCR), \emph{Multimedia Information Retrieval} (MIR). Each task makes an attempt to understand the correspondence between image and text. Almost all the V\&L framework is slightly different in feature processing mechanism. For example, ViLBERT \cite{lu2019vilBERT} propose a joint model for learning task-agnostic visual grounding from paired vision-linguistic data, where the Co-Attentional Transformer layers are designed to formulate the interaction between image and text. While VisualBERT\cite{li2019visualBERT} applies the serial Transformer layers and resets the pre-training task. In term of feature extraction, the same procedure is operated. Specifically, for visual representation, the image region features are extracted based on bounding boxes from the pre-trained objection detection network \cite{ren2015faster}. Then, a special \textbf{IMG} \emph{(for ViLBERT only)} token is assigned at the beginning of an image region to represent the whole image. For linguistic representation, the text is turned into discrete tokens, position encoding along with a small set of special tokens: \textbf{SEP, CLS, MASK}. However, the corresponding complementary information is missing during encoding the two different modality. Visual representation with only ROIs features lacks the discriminative contextual information which could connect in series things at the object level through contextual information effectively. Simultaneously, linguistic representation should be combined with more detailed variables (\emph{e.g.}, sensitive object-level information).

To further exploit information more comprehensively, this paper proposes a  Complementary Vision and Linguistic (CVL) networks based on the contextual-level and sensitive object-level information complementarity. The main contributions are three folds.
\begin{itemize}
\item   A Comprehensive Visual and Linguistic framework is designed for the Hateful Memes Challenge 2020.
\item   During training process, a sample augmentation method is proposed for different samples of the same label participate.
\item   Experimental results on Hateful Memes Challenge Dataset demonstrate that CVL provides a competitive performance on the criteria of AUROC and Accuracy.
\end{itemize}

\section{Related Work}
In this section, we brifely review the multi-modal sentiment analysis and visual and linguistic framework.

\subsection{Multi-modal Sentiment Analysis}
Multi-modal sentiment analysis is to predict the polarity of a group of image and text, which is similar to the hateful memes detection. In VisualNet \cite{truong2019vistanet}, the visual and textual components are leveraged, where images are utilized as an auxiliary part to prompt the salient content when dealing with text. In Multi-Interactive Memory Network (MIMN) \cite{xu2019multi}, the authors believe that the information contained in the corresponding aspect is critical for the extraction of text and image information, and design a multi-interactive aspect-guided attention mechanism to guide the model for generating attention vectors. In Hierarchical Fusion Model (HFM) \cite{cai2019multi}, the attribute of the image are imposed on the basis of the text and image modalities. Inspired by the relevant information supplemented papers and combined with actual work, we consider a more comprehensive information representation than previous methods, that is, information that combines contextual-level (\emph{global}) and sensitive object-level (\emph{local}) information. Both the global feature in visual and local feature in linguistic are considered in our model to build a complementary relational model.

\subsection{Visual and Linguistic Framework}
With the rapid development of Computer Vision\ (CV) and Natural Language Processing (NLP), visual-linguistic tasks are gradually emerging, giving birth to a series of visual-language general frameworks and the corresponding variants \cite{lu2019vilBERT,li2019visualBERT,sun2019videoBERT,tan2019lxmert,su2019vl}. In ViLBERT and VisualBERT, off-the-shelf image region features obtained from pre-trained object detection models and global text feature in BERT \cite{devlin2019BERT} manner are utilized to learning a optimal model. Both of above do not consider the discrimination of the complementary information. In VLBERT \cite{su2019vl}, for visual element corresponding to an region of interests \emph{(ROIs)}, the pre-trained detector is adopted to extract features, while the feature extraction is operated based on the whole input image for non-visual elements. PixelBERT \cite{huang2020pixel} aims to build a thorough connection between each image pixel and linguistic semantics directly from image and text pairs, instead of using region-based image features as that the feature representation capability is limited to the given categories in the specific dataset. In addition, in above mentioned framework, either a single-stream structure or a dual-stream structure has different emphasis on self-attention and co-attention. Given above issues, we design a complementary strategy of integrating the visual and linguistic framework to further exploit the relevant information, where both single-stream and dual-stream structure are contained to formulate the complementary information.

\begin{figure*}[!htb]
  \centering
  \includegraphics[width=0.85\textwidth{},keepaspectratio]{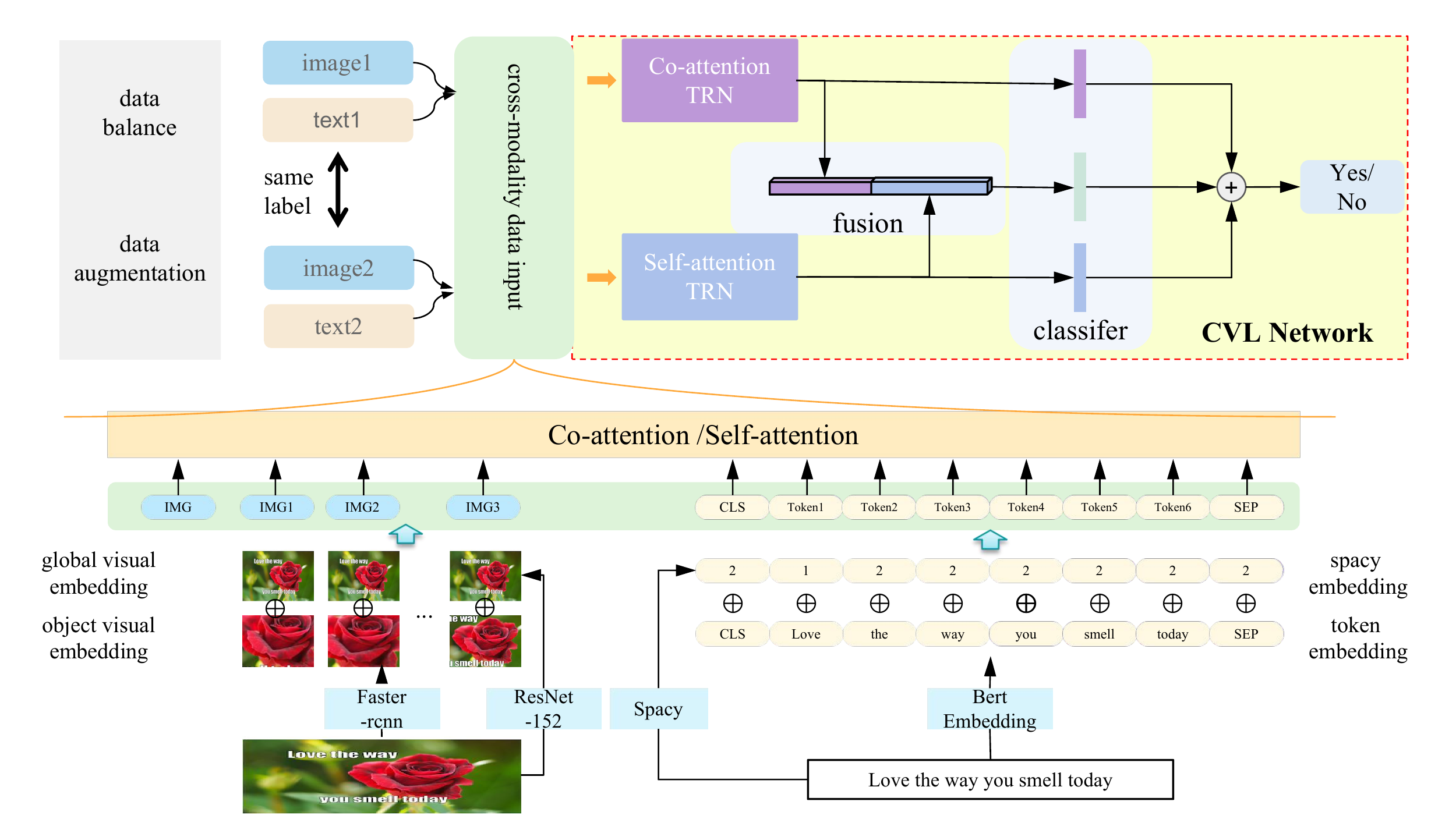}
  \caption{The overall framework of CVL network for hateful memes detection.}\label{fig:pipeline}
\end{figure*}

\section{Methodology}
\label{sec:method}
Figure \ref{fig:pipeline} illustrates the overall framework of the proposed CVL network. In the following subsections, we will describe image representation text representation and CVL network in detail.

\subsection{Image Representation}
We divide image representation into three components, namely object-level representation (\emph{i.e.}, visual elements corresponding to \emph{ROIs}), the \emph{ROIs} position embedding, and the whole contextual representation. In terms of object-level representation, we adopt Faster-RCNN detector pre-trained \cite{ren2015faster} on visual geometry dataset \cite{krishna2017visual} as the feature extractor to obtain each \emph{ROIs} representation. In terms of the whole contextual representation, we adopt ResNet-152 \cite{he2016deep} pre-trained on ImageNet \cite{deng2009imagenet} as the feature extractor. The feature is extracted from the last convolution layer ($2048$-Dim).
Like other visual-linguistic frameworks, the \emph{ROIs} position embedding is also considered. The 4-D representation, which denotes the geometry location (\emph{e.g.}, top-left and bottom-right corner, respectively) of the \emph{ROIs}, is integrated into the image representation.
The final image representation $V$ is formed through the fusion of three components in dot addition manner. Each feature of \emph{ROIs} are added the contextual representation and position embedding as follows:
\begin{equation}
V = V_{ROIs} \dotplus V_{contextual} \dotplus V_{pos},
\end{equation}
where $\dotplus$ denotes dot addition. $V_{ROIs}$ denotes the features of \emph{ROIs}, $V_{contextual}$ denotes the features of the whole image, $V_{pos}$ denotes the geometry location of \emph{ROIs}.

\subsection{Text Representation}
We observe that noun phases in the text are significant for representation. Therefore, different from existing works, besides the textual representation which follows the practice in BERT commonly used in visual-linguistic tasks, we design \textbf{Sembedding} symbol based on the open source tool Spacy\footnote{https://spacy.io/}. We first use Spacy to extract noun phases from text for each sample, and perform stopword filtering on the corresponding noun phases to obtain the S-keywords (\emph{Spacy keywork}) of each text. Then, we set a Sembedding symbol to indicate whether the each input token is a keyword. Symbol is set to $0, 2, 1$ to denotes that padding word, S-keyword and other sentence tokens, respectively. The final text representation is denoted as follows:
\begin{equation}
L = L_{BERT} \dotplus L_{Spacy},
\end{equation}
where $\dotplus$ denotes dot addition. $L_{BERT}$ denotes the text representation following BERT practice, $L_{Spacy}$ denotes Sembedding of corresponding text.

\subsection{CVL Network}
CVL network is composed of two basic visual-linguistic framework (namely ViLBERT and VisualBERT) with co-attention and self-attention, respectively. The former is a dual-stream architecture, where image and text are processed apart to provide a common representation. While another is a single-stream architecture. Given that the above two networks focus on different areas during the actual training process, we combine the two as the proposed CVL training network. In this paper, we simply concatenate the two networks before the final classification layer. Three fully-connected (\emph{FC} layers (no weights sharing) are applied to three-way features as classifier. Cross-entropy Loss is adopted as the final loss function.

\begin{table}[]
\centering
\begin{tabular}{cccc}
\toprule
\multirow{2}{*}{Types}                                                                          & \multirow{2}{*}{Models} & \multicolumn{2}{c}{Results (\%)}                    \\
                                                                                               &                        & Acc.                 & AUROC                \\
                                                                                               \hline
                                                                                               & Human                  & 84.70                & 82.65                \\
                                                                                               \hline
\multirow{3}{*}{Uni-modal}                                                                      & Image-Grid             & 52.00          & 52.63           \\
                                                                                               & Image-Region           & 52.13           & 55.92           \\
                                                                                               & Text BERT              & 59.20           & 65.08           \\
                                                \hline
\multirow{6}{*}{\begin{tabular}[c]{@{}c@{}}Multi-modal\\ (Uni-modal Pre-training)\end{tabular}}   & Late Fusion            & 59.66           & 64.75           \\
                                                                                               & Concat BERT            & 59.13           & 65.79          \\
                                                                                               & MMBT-Grid              & 60.06           & 67.92           \\
                                                                                               & MMBT-Region            & 60.23           & 70.73           \\
                                                                                               &  ViLBERT                & 62.30           & 70.45           \\
                                                                                               &  Visual BERT            & 63.20           & 71.33          \\
                                                                                               \hline
\multirow{2}{*}{\begin{tabular}[c]{@{}c@{}}Multi-modal\\ (Multi-modal Pre-training)\end{tabular}} & ViLBERT CC             & 61.10           & 70.03           \\
                                                                                               & Visual BERT COCO       & 64.73           & 71.41           \\
                                                                                               \hline
                                                                                               \hline
\multirow{3}{*}{\begin{tabular}[c]{@{}c@{}}Multi-modal\\ (Uni-modal Pre-training)\end{tabular}}                                                                           &  ViLBERT with Contextual and Sembedding                   & \underline{65.30} & \textbf{75.22}\\
\multicolumn{1}{l}{}                                                                           &  VisualBERT with Contextual and Sembedding                   & 64.30 & 72.85\\
\multicolumn{1}{l}{}                                                                           & CVL  & \textbf{66.20} & \underline{75.02}       \\
\bottomrule
\end{tabular}
\caption{Experimental results comparison of the proposed CVL with other methods. \textbf{Bold} denotes the best and the \underline{underline} denotes the suboptimal.}
\label{ref:result}
\end{table}
\section{Experiment}
\subsection{Dataset}
\paragraph{Hateful Memes Challenge Dataset} The dataset\cite{kiela2020hateful} is released by Facebook AI for Hateful Memes Challenge 2020, which totally consists of $10K$ memes, including five different types, namely multi-modal hate, where benign confounders were found for both modalities, uni-modal hate where one or both modalities were already hateful on their own, benign image and benign text confounders and finally random not-hateful examples. There are $8.5K$, $0.5K$ memes and $1K$ memes for training, validation and testing, respectively. When the Challenge progressed to the Stage-$2$, the labels of some samples in the training set are revised and dev\_unseen is new released. Therefore, we update the labels, add dev\_seen to the training set, and use dev\_unseen as the validation set.

\subsection{Implementation Details}
We follow the multi-modal framework (MMF)\footnote{https://github.com/facebookresearch/mmf} settings. For image representation, Faster-RCNN detector pre-trained on Visual Genome \cite{krishna2017visual} is utilized to extract $100$ \emph{ROIs}, including the corresponding \emph{ROIs} features, bounding boxes. The batch size is set to $80$. The initial learning rate is set $1e-5$ and $5e-5$ for ViLBERT and VisualBERT respectively with warm-up before $2,000$ iteration. Adam \cite{kingma2014adam} is adopted as the optimizer.
The basic setting follow MMF. While the total iteration is set $22,000$. For Sembedding in text representation, each text has the corresponding Vocabulary (no Sembedding sharing).
\subsection{Experimental Results}
Experimental results comparison of the proposed CVL with other methods are shown in Table \ref{ref:result}. We conduct the contextual and Sembedding experiment with ViLBERT, VisualBERT, CVL Network. Compared to Multi-modal (Uni-modal Pre-training), as is illustrated, ViLBERT gets a promotion of $+3.0\%$ and $+4.77\%$ on criteria of Accuracy and AUROC. VisualBERT increases by $+1.1\%$ for Accuracy and $+1.52\%$ for AUROC, respectively. CVL achieves $66.2\%$ for Accuracy and $75.02$ for AUROC. Compared to VilBERT with contextual and Sembedding , CVL provides a more excellent Accuray. We argue that the reason maybe that the combination of self-attention and mutual attention leads to the model understanding of some examples during learning. Fusion by average decision with 20 other models which are trained with uni-modal or multi-modal, our CVL method achieves $78.48\%$ and $72.95$\% on the criteria of AUROC and Accuracy on Stage-\emph{2} on Hateful Memes Challenge 2020.

\section{Conclusion}
This paper investigates a Complementary Visual and Linguistic (CVL) network in Hateful Memes Challenge 2020 to ensure that the more comprehensive information of the multi-modality could be explored in detail. With the contextual-level and sensitive object-level information in visual and linguistic embedding, the proposed CVL network could provide a solution candidate for the complex multi-modal scenarios.
\bibliographystyle{unsrt}


\bibliography{references}
\end{document}